\documentclass[conference]{IEEEtran}
\IEEEoverridecommandlockouts
\usepackage{cite}
\usepackage{amsmath,amssymb,amsfonts}
\usepackage{algorithmic}
\usepackage{graphicx}
\usepackage{textcomp}
\usepackage{xcolor}
\usepackage{soul}
\def\BibTeX{{\rm B\kern-.05em{\sc i\kern-.025em b}\kern-.08em
    T\kern-.1667em\lower.7ex\hbox{E}\kern-.125emX}}
\usepackage{booktabs,dcolumn}

\begin{document}

\title{Detection of Adversarial Attacks on Super-Resolvers Using Spectral Features\\
\thanks{This manuscript has been authored in part by UT-Battelle, LLC, under contract DE-AC05-00OR22725
with the US Department of Energy (DOE). The publisher acknowledges the US government license to provide public access under the DOE Public Access Plan (https://energy.gov/doe-publicaccess-plan). This research was sponsored by the Laboratory Directed Research and Development Program of Oak Ridge National Laboratory, managed by UT-Battelle, LLC, for the U. S. Department of Energy. This research used resources from the ORNL Research Cloud Infrastructure at the Oak Ridge National Laboratory, which is supported by the Office of Science of the U.S. Department of Energy under Contract No. DE-AC05-00OR22725.}
}

\author{\IEEEauthorblockN{Emma J. Reid}
\IEEEauthorblockA{\textit{Human Analysis and Biometrics} \\
\textit{Oak Ridge Laboratory}\\
Oak Ridge, TN, USA \\
reidej@ornl.gov}
\and
\IEEEauthorblockN{Haley Duba-Sullivan}
\IEEEauthorblockA{\textit{Radar and Computational Imaging} \\
\textit{Oak Ridge Laboratory}\\
Oak Ridge, TN, USA \\
sullivanhe@ornl.gov}
\and
\IEEEauthorblockN{Tony G. Allen}
\IEEEauthorblockA{\textit{Radar and Computational Imaging} \\
\textit{Oak Ridge Laboratory}\\
Oak Ridge, TN, USA \\
allentg@ornl.gov}
}

\maketitle

\begin{abstract}
The integration of deep learning models into image preprocessing pipelines such as super-resolution introduces a largely unexplored attack vector for adversaries targeting downstream tasks.
To ensure trustworthiness of critical imaging pipelines, we must be able to detect adversarial behavior within preprocessing models. 
In this paper, we propose a spectral-based detection method for identifying adversarial attacks embedded in super-resolution model weights.
More specifically, we use the radially-averaged power spectral density as a discriminative feature to train an extreme gradient boosting (XGBoost) detector, demonstrating detectability of model-level threats in super-resolution networks. 
We further benchmark our detector against magnitude- and phase-based Fourier spectrum detectors, evaluating each method across a range of training and cross-architecture scenarios.
Our proposed detector out-performs the comparison detectors in most of these scenarios and indicates that high-frequency features are most informative for detecting AdvSR attacks across SR architectures.

\end{abstract}


\section{Introduction}
Preprocessing pipelines are essential for state-of-the-art quality in downstream computer vision tasks, such as detection and classification \cite{CatarinoSegmentation, SchegolikhinObject, ZhouClassification}. However, the incorporation of deep learning models into preprocessing pipelines introduces a stealthy attack vector for adversaries targeting downstream tasks. To ensure trustworthiness of imaging pipelines, it is essential to design methods to detect adversarial behavior in preprocessing models. 

\begin{figure}
    \centering
    \includegraphics[width=0.95\linewidth]{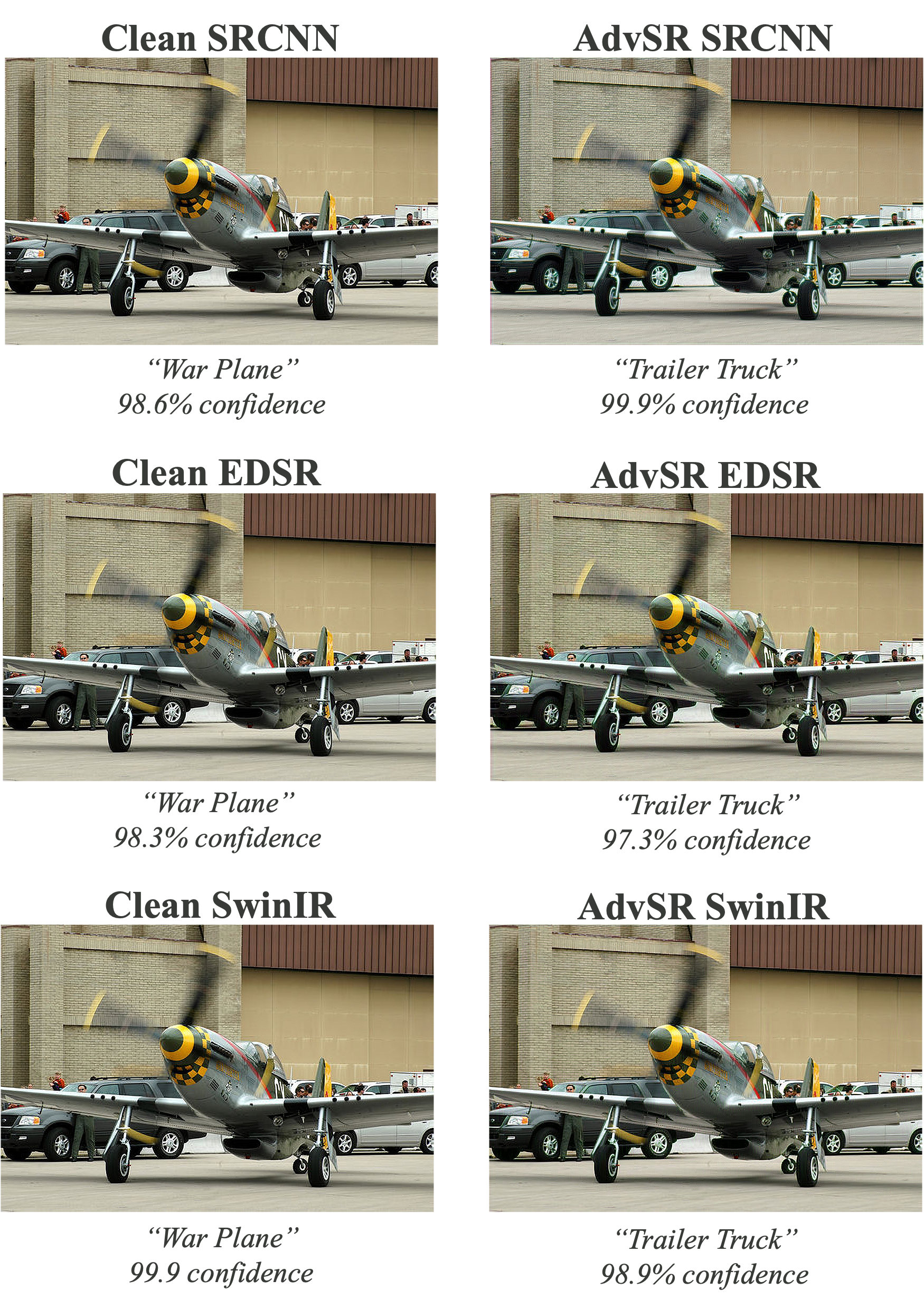}
    \caption{Example of AdvSR attacks for SRCNN, EDSR, and SwinIR architectures targeting misclassification of ``war planes'' to ``trailer trucks'' for YOLOv11. These attacks preserve image quality while causing misclassification with high confidence.}
    \label{fig:ex-attack}
\end{figure}

In our previous work, we constructed a targeted adversarial attack embedded within the weights of a super-resolution model \cite{FY25_report, dubasullivan2026-sword}. The resulting adversarial super-resolution (AdvSR) models produce high-fidelity reconstructions, while causing targeted misclassification. We showed successful attacks using three distinct super-resolution architectures: SRCNN \cite{SRCNN}, EDSR \cite{EDSR}, and SwinIR \cite{SwinIR}. 
Since then, we refined the attack to be effective for classifiers with more classes, less perceptible to the human eye, and adaptable to any input image size \cite{FY26_report}. Figure~\ref{fig:ex-attack} provides an example of the improved AdvSR attack targeting misclassification of ``war planes'' to ``trailer trucks'' for YOLOv11 \cite{redmon2016lookonceunifiedrealtime, yolo11_ultralytics}.
While our previous work explored AdvSR's viability as an attack vector, we have yet to explore the detectability of this attack. 

In this paper, we propose a novel spectral-based detector and evaluate its performance on AdvSR attacks.
This detector exploits an image's radially-averaged power spectral density (PSD) as a discriminative feature for detecting AdvSR attacks. We compare our approach to existing spectral approaches and demonstrate better performance in the majority of test cases.
We also examine the explainability of our proposed method and identify that AdvSR attacks inject most adversarial content into high spatial frequencies.

\section{Related Work}

Detection methods for adversarial attacks apply a variety of techniques to detect malicious images from clean images.
Local Intrinsic Dimensionality (LID) \cite{ma2018characterizingadversarialsubspacesusing} focuses on identifying adversarial subspaces through increased dimensionality, while deep Mahalanobis detectors \cite{LeeMahalanobis2018} employ the Mahalanobis distance to identify out-of-distribution samples. 
Squeezing methods like \cite{RyuDetectionEntropy, XuFeatureSqueezing2017} compress image feature spaces and use the level of compression to detect attacks.
Among spectral approaches, SpectralDefense \cite{HarderCNNSFourier} introduces four frequency-domain detectors. These detectors apply logistic regression to the magnitude or phase of the Fourier spectrum computed from either the input image or internal layer activations of the classification model. Collectively, these methods demonstrate state-of-the-art performance across five standard attacks, including FGSM \cite{fgsm2015} and Carlini \& Wagner \cite{carlini2017evaluatingrobustnessneuralnetworks}. However, these standard attacks have been shown to alter features in medium frequency bands~\cite{yin2019}, while super-resolution methods alter high-frequency content by design. Thus a successful AdvSR detector must distinguish between genuine and malicious high-frequency content.

Moreover, many of the aforementioned detectors require white-box access to the classifier model to function. This limits their applicability in practice, where this access may be unavailable.
Instead, our proposed method uses only spectral features of the SR image and does not require access to the downstream image classifier, its weights, or its internal activations.
We compare our proposed approach with two other image-only detectors, InputMFS and InputPFS \cite{HarderCNNSFourier}, which generally outperformed LID and Mahalanobis distance.
This enables a direct comparison of image-level spectral features for detecting AdvSR attacks.

\section{Proposed Method}

For each super-resolved image, we first compute a single-channel luminance image, $x$, and then calculate its 2D Fourier transform, shifted so that the zero-frequency component lies at the center. Let $X(u,v)$ be the resulting spectrum and $P(u,v) = |X(u,v)|^2$ its power spectrum.
Next, we group the spectral coefficients into bins based on their normalized radial distance from the center and average the coefficients within each bin.
The resulting radially-averaged power spectral density (PSD) for the $k$th bin spanning radial distance $[r_k, r_{k+1}) \subseteq [0,1]$ is given by
\begin{equation}
    \operatorname{PSD}(r_k) = \frac{1}{|S_k|} \sum_{(u,v)\in S_k} P(u,v) 
\end{equation}
where $S_k$ is the set of frequencies $(u,v)$ in the $k$th bin and $|S_k|$ denotes the number of elements in that set.

We use $\operatorname{PSD}(r_k)$ with 256 frequency bins as an input feature vector for our detector.
By inputting the radially-averaged PSD, we ensure that all frequencies across the image spectrum are included for analysis while also compressing features through radial averaging.
We provide this vector to an extreme gradient boosting (XGBoost) \cite{XGBoost} classifier, which predicts whether the image was generated by a clean or adversarial super-resolution model.
We select XGBoost for its speed and innate explainability. 

\section{Experimental Data and Results}
In this section, we detail training procedures, report experimental results, and discuss explainability of our proposed method and its implications.

\begin{figure*}
    \centering
\includegraphics[width=0.95\linewidth]{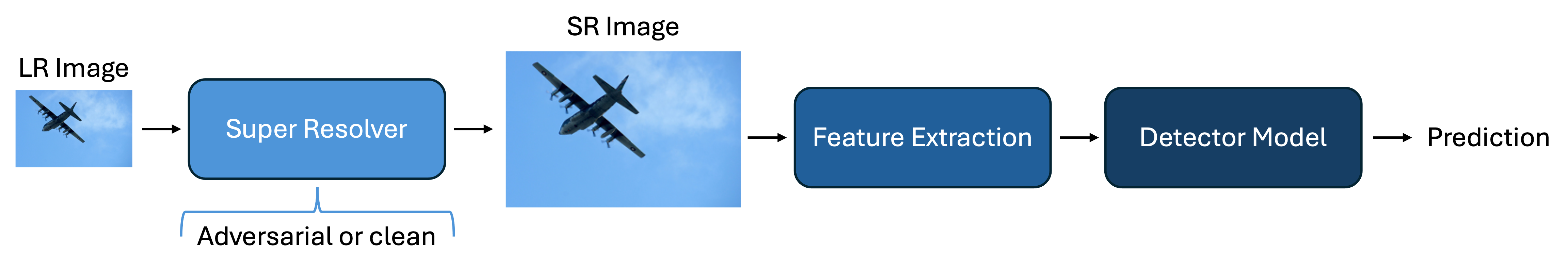}
    \caption{Pipeline for training and testing detection methods. We apply clean and adversarial super-resolvers to LR images, apply feature extractors to the resulting SR images, and input these features to detectors for both training and testing.}
    \label{fig:pipeline}
\end{figure*}

\subsection{Training Details}

For training and testing, we use a subset of ImageNet \cite{deng2009imagenet} containing vehicular classes such as sports car, tractor trailer, and war plane, as well as three non-vehicular dummy classes.
To assess the impact of classifier complexity on attack detectability, we construct a dataset from this subset containing 5 classes (ImageNet5). 
These images are then split into training, validation, and testing sets with distinct training sets for the super-resolvers, YOLOv11 classifier, and XGBoost detector.

Figure~\ref{fig:pipeline} shows the procedure used for detector training and testing. We input LR images into both clean and AdvSR models to produce SR images. These are then labeled as 0 for clean and 1 for adversarial. All detectors (including comparison methods) receive the same clean and adversarial training images. Then we extract spectral features from the super-resolved images, which are the radially-averaged PSD for our approach and magnitude and/or phase of the Fourier spectrum for InputMFS and InputPFS. Finally, we use the extracted features to train the detectors. 

\subsection{Detector Results}
We train individual detectors on SRCNN, EDSR, and SwinIR architectures respectively.
Additionally, we jointly train on two architectures and test on the unseen architecture (``leave-one-out''), as well as train a unified detector on all three architectures. We evaluate all detectors on the same testing images at inference time.
Unless specified, we test all detectors on their trained AdvSR architectures. 
We assess performance across all detectors using accuracy (Acc), F1 score (F1) and area under the curve (AUC). 

\begin{table*}[t]
    \centering
    \caption{Detector performance across SR architectures. Each detector is trained and tested using outputs from the same SR architecture. The best result for each architecture and metric is shown in bold.}
    \scriptsize
    \setlength{\tabcolsep}{5pt}
    \begin{tabular}{lccccccccc}
        \toprule
        & \multicolumn{3}{c}{Ours} & \multicolumn{3}{c}{InputMFS} & \multicolumn{3}{c}{InputPFS} \\
        \cmidrule(lr){2-4} \cmidrule(lr){5-7} \cmidrule(lr){8-10}
        \textbf{SR model} & \textbf{Acc.} & \textbf{F1} & \textbf{AUC} & \textbf{Acc.} & \textbf{F1} & \textbf{AUC} & \textbf{Acc.} & \textbf{F1} & \textbf{AUC} \\
        \midrule
        SRCNN & 0.845 & 0.840 & 0.926 & \textbf{0.890} & \textbf{0.888} & \textbf{0.954} & 0.481 & 0.445 & 0.498 \\
        EDSR   & \textbf{0.970} & \textbf{0.969} & \textbf{0.997} & 0.886 & 0.887 & 0.952 & 0.485 & 0.456 & 0.488 \\
        SwinIR & \textbf{0.962} & \textbf{0.962} & \textbf{0.997} & 0.917 & 0.919 & 0.978 & 0.523 & 0.526 & 0.524 \\
        \bottomrule
    \end{tabular}
    \label{tab:detector-comparison}
\end{table*}

In Table~\ref{tab:detector-comparison}, we directly compare the proposed radially-averaged PSD detector with InputMFS and InputPFS when the detectors are trained and tested on the same SR architecture. Our proposed detector achieves the highest accuracy, F1 score, and ROC-AUC for EDSR and SwinIR, while InputMFS performs best for SRCNN. Averaged across the three architectures, the proposed detector obtains an F1 score of 0.924 and ROC-AUC of 0.973, compared with 0.898 and 0.962 for InputMFS and 0.476 and 0.503 for InputPFS, respectively. 
These results indicate that the radially-averaged PSD is a better discriminative feature for the AdvSR attack using complex SR architectures.

\begin{table*}[t]
    \centering
    \caption{Detector performance under leave-one-SR-model-out and unified training. The unified detectors are trained using all three SR architectures; ``All SR'' is their pooled test set. The best result for each test set and metric is shown in bold.}
    \scriptsize
    \setlength{\tabcolsep}{3.2pt}
    \begin{tabular}{llccccccccc}
        \toprule
        & & \multicolumn{3}{c}{Ours} & \multicolumn{3}{c}{InputMFS} & \multicolumn{3}{c}{InputPFS} \\
        \cmidrule(lr){3-5} \cmidrule(lr){6-8} \cmidrule(lr){9-11}
        \textbf{Training} & \textbf{Test SR} & \textbf{Acc.} & \textbf{F1} & \textbf{AUC} & \textbf{Acc.} & \textbf{F1} & \textbf{AUC} & \textbf{Acc.} & \textbf{F1} & \textbf{AUC} \\
        \midrule
        Leave-one-out & SRCNN & 0.500 & 0.000 & 0.564 & \textbf{0.576} & 0.446 & \textbf{0.649} & 0.462 & \textbf{0.462} & 0.472 \\
        Leave-one-out & EDSR   & \textbf{0.966} & \textbf{0.967} & \textbf{0.999} & 0.723 & 0.700 & 0.788 & 0.470 & 0.496 & 0.481 \\
        Leave-one-out & SwinIR & \textbf{0.864} & \textbf{0.846} & \textbf{0.936} & 0.769 & 0.775 & 0.844 & 0.504 & 0.513 & 0.525 \\
        \midrule
        Unified & SRCNN & \textbf{0.856} & 0.854 & \textbf{0.934} & 0.848 & \textbf{0.856} & 0.921 & 0.511 & 0.524 & 0.491 \\
        Unified & EDSR   & \textbf{0.985} & \textbf{0.985} & \textbf{1.000} & 0.818 & 0.824 & 0.886 & 0.492 & 0.538 & 0.487 \\
        Unified & SwinIR & \textbf{0.962} & \textbf{0.962} & \textbf{0.995} & 0.860 & 0.865 & 0.922 & 0.530 & 0.572 & 0.544 \\
        Unified & All SR & \textbf{0.934} & \textbf{0.934} & \textbf{0.984} & 0.842 & 0.848 & 0.910 & 0.511 & 0.545 & 0.508 \\
        \bottomrule
    \end{tabular}
    \label{tab:generalization}
\end{table*}

In Table~\ref{tab:generalization}, we show the results for leave-one-out and unified approaches. On average, the proposed detector obtains a ROC-AUC of 0.833, compared with 0.760 for InputMFS and 0.493 for InputPFS. Its performance is strongest when EDSR or SwinIR is held out. However, when SRCNN is held out, InputMFS performs best and the proposed detector classifies every adversarial SRCNN output as clean.  
We also assess the performance of a unified model for all detectors with performance similar to Table~\ref{tab:detector-comparison} across all detectors. The unified detector results on solely SRCNN test images suggest that SRCNN is not a consistent failure case for all models and rather reflects a correctable lapse in training data. As expected, the inclusion of more SR architectures and their spectral behavior in training improves the generalization of the detector. This discrepancy in performance across SR architectures motivates exploring the proposed detector's explainability.

\subsection{Detector Explainability}

\begin{figure*}[t]
    \centering
    \includegraphics[width=0.95\textwidth]{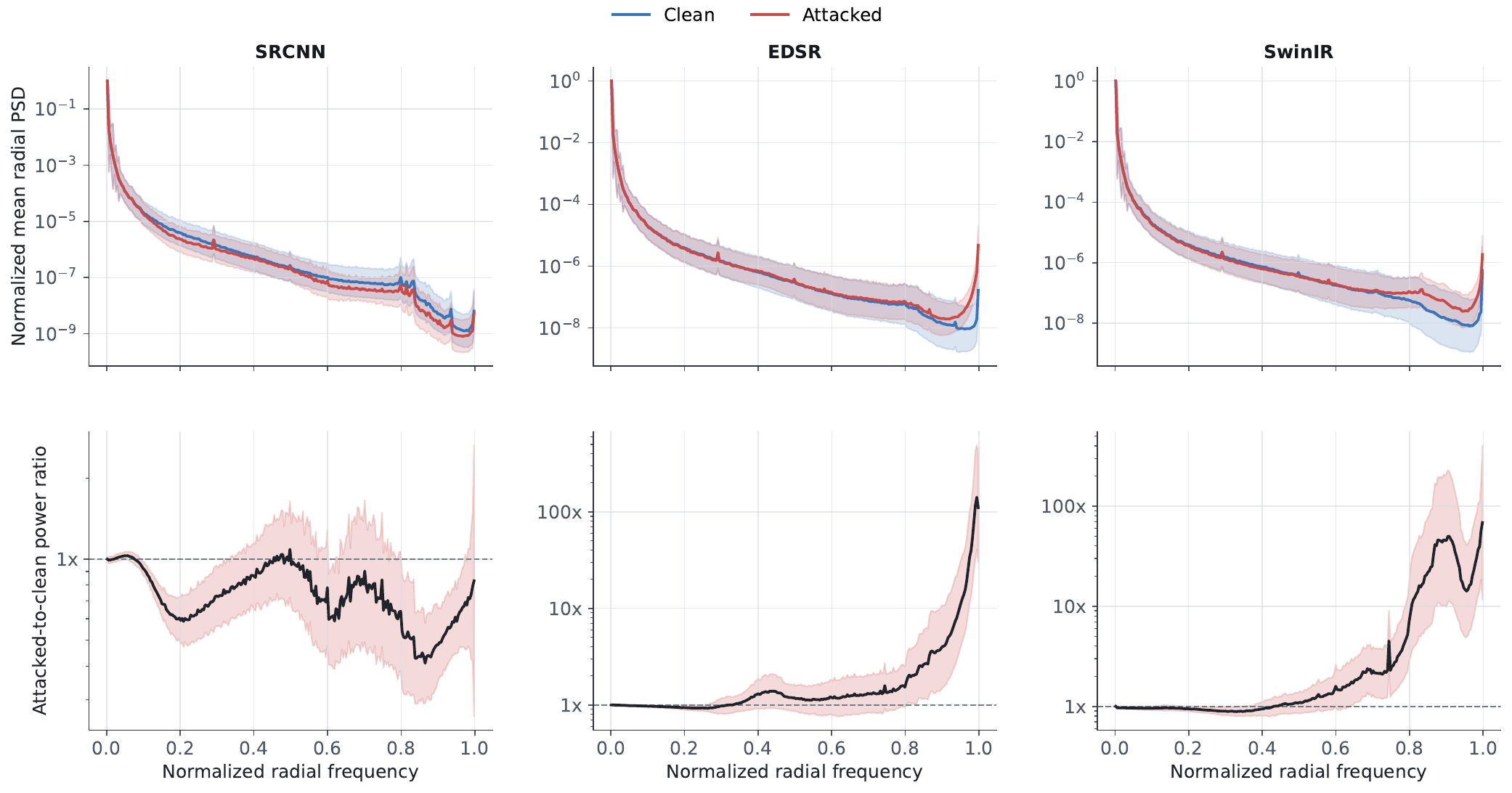}
    \caption{Normalized radially-averaged PSD for clean and adversarial SR outputs. Each spectrum is normalized by its DC bin before averaging. Solid curves show the mean, while the shaded regions indicate one standard deviation above and below the mean. The lower row shows the attacked-to-clean power ratio, where a ratio of one indicates no change.}
    \label{fig:radial-psd}
\end{figure*}

In Figure~\ref{fig:radial-psd}, we compare the normalized mean radially-averaged PSD of clean and adversarial outputs across 256 frequency bins for each SR architecture. The adversarial EDSR and SwinIR outputs exhibit a large increase in power at the highest radial frequencies. In contrast, the adversarial SRCNN outputs are actually lower than the clean outputs at most of these frequencies. 
This power mismatch between SRCNN and the other two examined architectures helps explain the discrepancies seen in Table~\ref{tab:generalization}, as the attacked-to-clean power ratio increases sharply for EDSR and SwinIR. Without this increase in power, a detector trained on these architectures could reasonably infer that the input image was clean. 

\begin{figure*}
    \centering
    \includegraphics[width=\textwidth]{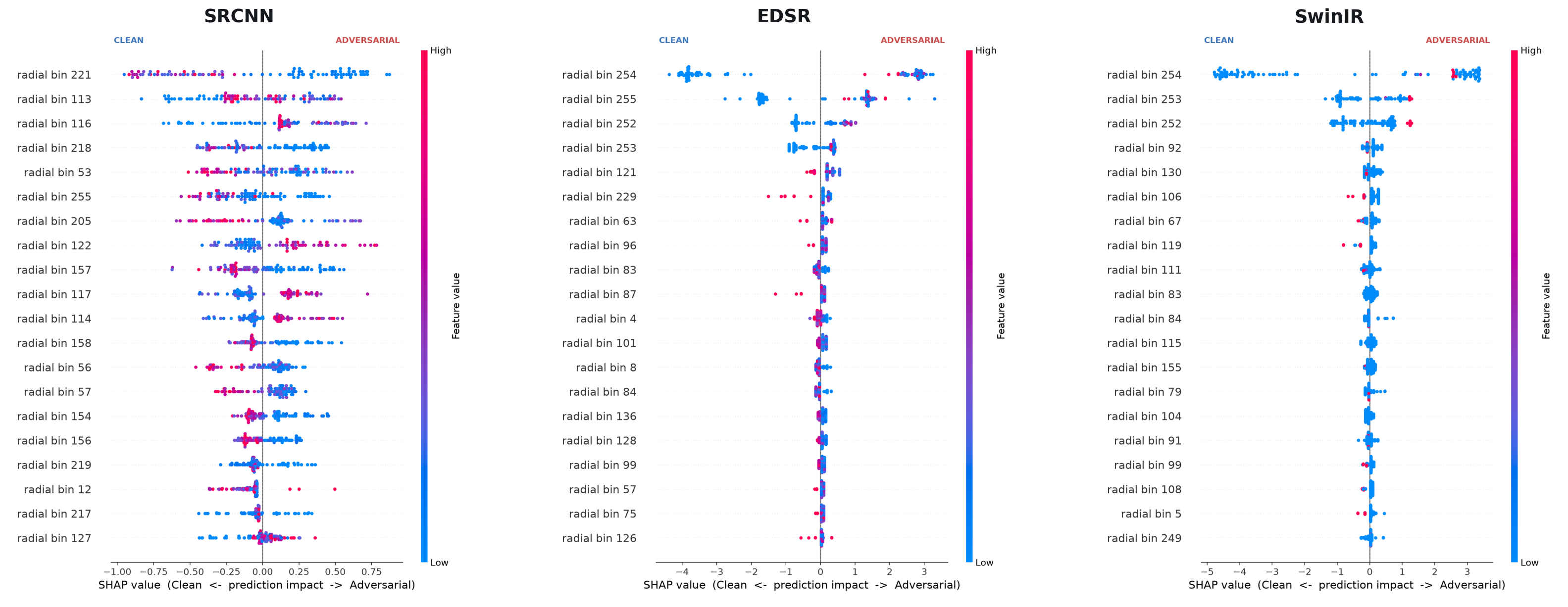}
    \caption{Shapley analysis of the proposed radially-averaged PSD detector across SR architectures. Feature color represents radial spectral power, while the horizontal Shapley value represents the contribution toward a clean or adversarial prediction.}
    \label{fig:SHAP}
\end{figure*}

In Figure~\ref{fig:SHAP}, we further examine the frequency bins that drive the proposed detector's predictions for each SR architecture, using Shapley values \cite{lundberg2017unifiedapproachinterpretingmodel}. Positive Shapley values move a prediction toward the adversarial class, while negative values move it toward the clean class; the feature value indicates the radial spectral power observed in that bin. For EDSR and SwinIR, the largest absolute Shapley values are concentrated in the highest-frequency bins, indicating that the detector identifies these attacks primarily from changes near the upper end of the radial spectrum. For SRCNN, however, the influential features are distributed across a broader range of frequencies, suggesting that its attack modifies the spectral signature more diffusely. This aligns with the radially-averaged PSD results shown in Figure~\ref{fig:radial-psd}. Notably, the results from AdvSR attacks on EDSR and SwinIR contrast with traditional adversarial attacks, such as Carlini\&Wagner, which have been observed to predominately feature in mid-frequency bands~\cite{yin2019, HarderCNNSFourier}. A detector that evaluates features across low, mid, and high frequencies would likely then be the most robust to changes in SR models and attacks.

\section{Conclusion}

In this work, we demonstrate the detectability of AdvSR attacks using spectral features. Our developed detector generalizes across AdvSR models and produces superior metrics to comparison detectors in most testing scenarios. We additionally incorporate explainability to investigate where our detector falls short and why. Future work will use these results to improve the robustness of our detector and increase its generalizability to other SR architectures and model-weight attacks.

\bibliographystyle{IEEEtran}
\bibliography{AdvSR}

\end{document}